\documentclass[11pt]{article}
\usepackage{acl2016}
\usepackage{url}
\usepackage{times}
\usepackage{latexsym}
\aclfinalcopy 

\usepackage[utf8]{inputenc}
\usepackage[T1]{fontenc}

\usepackage{CJKutf8}

\usepackage[russian,english]{babel}

\usepackage{enumitem}
\usepackage{amsmath}
\usepackage{multirow}
\usepackage{tikz-qtree}
\usepackage{tikz-dependency}
\usepackage{adjustbox}

\usepackage{pgfplots}
\pgfplotsset{compat=newest}

\usepackage{siunitx}
\newcommand{\pipe}{{\color{gray}|}}

\usepackage{textcomp}
\usepackage{algorithm}
\usepackage{listings}{
\lstset{
  language=Python,
  upquote=true,
  showstringspaces=false,
  formfeed=newpage,
  tabsize=4,
  commentstyle=itshape,
  basicstyle=\tiny\ttfamily,
  morekeywords={models, lambda, forms}
}

\setlength\titlebox{5cm}

\renewcommand\footnotemark{}
\title{Improving Neural Machine Translation Models with Monolingual Data\thanks{The research presented in this publication was conducted in cooperation with Samsung Electronics Polska sp.\ z o.o.\ - Samsung R\&D Institute Poland.} }

\author{
Rico Sennrich\and Barry Haddow \and Alexandra Birch\\
School of Informatics, University of Edinburgh\\
{\tt \{rico.sennrich,a.birch\}@ed.ac.uk}, {\tt bhaddow@inf.ed.ac.uk}
}

\date{}

\begin{document}
\maketitle
\begin{abstract}
Neural Machine Translation (NMT) has obtained state-of-the art performance for several language pairs, while only using parallel data for training.
Target-side monolingual data plays an important role in boosting fluency for phrase-based statistical machine translation, and we investigate the use of monolingual data for NMT.
In contrast to previous work, which combines NMT models with separately trained language models,
we note that encoder-decoder NMT architectures already have the capacity to learn the same information as a language model,
and we explore strategies to train with monolingual data without changing the neural network architecture.
By pairing monolingual training data with an automatic back-translation, we can treat it as additional parallel training data,
and we obtain substantial improvements on the WMT 15 task English$\leftrightarrow$German (+2.8--3.7 {\sc Bleu}), and for the low-resourced IWSLT 14 task Turkish$\to$English (+2.1--3.4 {\sc Bleu}), obtaining new state-of-the-art results.
We also show that fine-tuning on in-domain monolingual and parallel data gives substantial improvements for the IWSLT 15 task English$\to$German.
\end{abstract}

\section{Introduction}

Neural Machine Translation (NMT) has obtained state-of-the art performance for several language pairs, while only using parallel data for training.
Target-side monolingual data plays an important role in boosting fluency for phrase-based statistical machine translation, and we investigate the use of monolingual data for NMT.

Language models trained on monolingual data have played a central role in statistical machine translation since the first IBM models \cite{brown1990}.
There are two major reasons for their importance.
Firstly, word-based and phrase-based translation models make strong independence assumptions, with the probability of translation units estimated independently from context,
and language models, by making different independence assumptions, can model how well these translation units fit together.
Secondly, the amount of available monolingual data in the target language typically far exceeds the amount of parallel data,
and models typically improve when trained on more data, or data more similar to the translation task.

In (attentional) encoder-decoder architectures for neural machine translation \cite{DBLP:conf/nips/SutskeverVL14,DBLP:journals/corr/BahdanauCB14},
the decoder is essentially an RNN language model that is also conditioned on source context, so the first rationale, adding a language model to compensate for the independence assumptions of the translation model, does not apply.
However, the data argument is still valid in NMT, and we expect monolingual data to be especially helpful if parallel data is sparse, or a poor fit for the translation task, for instance because of a domain mismatch.

In contrast to previous work, which integrates a separately trained RNN language model into the NMT model \cite{DBLP:journals/corr/GulcehreFXCBLBS15},
we explore strategies to include monolingual training data in the training process without changing the neural network architecture.
This makes our approach applicable to different NMT architectures.

The main contributions of this paper are as follows:

\begin{itemize}
\item we show that we can improve the machine translation quality of NMT systems by mixing monolingual target sentences into the training set.
\item we investigate two different methods to fill the source side of monolingual training instances: using a dummy source sentence, and using a source sentence obtained via back-translation, which we call synthetic.
We find that the latter is more effective.
\item we successfully adapt NMT models to a new domain by fine-tuning with either monolingual or parallel in-domain data.
\end{itemize}

\section{Neural Machine Translation}

We follow the neural machine translation architecture by \newcite{DBLP:journals/corr/BahdanauCB14}, which we will briefly summarize here.
However, we note that our approach is not specific to this architecture.

The neural machine translation system is implemented as an encoder-decoder network with recurrent neural networks.

The encoder is a bidirectional neural network with gated recurrent units \cite{cho-EtAl:2014:EMNLP2014} that reads an input sequence $x=(x_1,...,x_m)$ and calculates a forward sequence of hidden states $(\overrightarrow{h}_1,...,\overrightarrow{h}_m)$,
and a backward sequence $(\overleftarrow{h}_1,...,\overleftarrow{h}_m)$.
The hidden states $\overrightarrow{h}_j$ and $\overleftarrow{h}_j$ are concatenated to obtain the annotation vector $h_j$.

The decoder is a recurrent neural network that predicts a target sequence $y=(y_1,...,y_n)$.
Each word $y_i$ is predicted based on a recurrent hidden state $s_i$, the previously predicted word $y_{i-1}$, and a context vector $c_i$.
$c_i$ is computed as a weighted sum of the annotations $h_j$.
The weight of each annotation $h_j$ is computed through an \emph{alignment model} $\alpha_{ij}$, which models the probability that $y_i$ is aligned to $x_j$.
The alignment model is a single-layer feedforward neural network that is learned jointly with the rest of the network through backpropagation.

A detailed description can be found in \cite{DBLP:journals/corr/BahdanauCB14}.
Training is performed on a parallel corpus with stochastic gradient descent.
For translation, a beam search with small beam size is employed.

\section{NMT Training with Monolingual Training Data}

In machine translation, more monolingual data (or monolingual data more similar to the test set) serves to improve the estimate of the prior probability $p(T)$ of the target sentence $T$, before taking the source sentence $S$ into account.
In contrast to \cite{DBLP:journals/corr/GulcehreFXCBLBS15}, who train separate language models on monolingual training data and incorporate them into the neural network through shallow or deep fusion,
we propose techniques to train the main NMT model with monolingual data, exploiting the fact that encoder-decoder neural networks already condition the probability distribution of the next target word on the previous target words.
We describe two strategies to do this:
providing monolingual training examples with an empty (or dummy) source sentence,
or providing monolingual training data with a synthetic source sentence that is obtained from automatically translating the target sentence into the source language,
which we will refer to as \textit{back-translation}.

\subsection{Dummy Source Sentences}

The first technique we employ is to treat monolingual training examples as parallel examples with empty source side, essentially adding training examples whose context vector $c_i$ is uninformative, and for which the network has to fully rely on the previous target words for its prediction.
This could be conceived as a form of dropout \cite{DBLP:journals/corr/abs-1207-0580}, with the difference that the training instances that have the context vector dropped out constitute novel training data.
We can also conceive of this setup as multi-task learning, with the two tasks being translation when the source is known, and language modelling when it is unknown.

During training, we use both parallel and monolingual training examples in the ratio 1-to-1, and randomly shuffle them.
We define an epoch as one iteration through the parallel data set, and resample from the monolingual data set for every epoch.
We pair monolingual sentences with a single-word dummy source side \textit{<null>} to allow processing of both parallel and monolingual training examples with the same network graph.\footnote{One could force the context vector $c_i$ to be 0 for monolingual training instances, but we found that this does not solve the main problem with this approach, discussed below.}
For monolingual minibatches\footnote{For efficiency, \newcite{DBLP:journals/corr/BahdanauCB14} sort sets of 20 minibatches according to length.
This also groups monolingual training instances together.}, we freeze the network parameters of the encoder and the attention model.

One problem with this integration of monolingual data is that we cannot arbitrarily increase the ratio of monolingual training instances, or fine-tune a model with only monolingual training data,
because different output layer parameters are optimal for the two tasks, and the network `unlearns' its conditioning on the source context if the ratio of monolingual training instances is too high.

\subsection{Synthetic Source Sentences}

To ensure that the output layer remains sensitive to the source context, and that good parameters are not unlearned from monolingual data, we propose to pair monolingual training instances with a synthetic source sentence from which a context vector can be approximated.
We obtain these through back-translation, i.e.\ an automatic translation of the monolingual target text into the source language.

During training, we mix synthetic parallel text into the original (human-translated) parallel text and do not distinguish between the two: no network parameters are frozen.
Importantly, only the source side of these additional training examples is synthetic, and the target side comes from the monolingual corpus.

\section{Evaluation}

We evaluate NMT training on parallel text, and with additional monolingual data, on English$\leftrightarrow$German and Turkish$\to$English,
using training and test data from WMT 15 for English$\leftrightarrow$German, IWSLT 15 for English$\to$German, and IWSLT 14 for Turkish$\to$English.

\subsection{Data and Methods}

We use Groundhog\footnote{\url{github.com/sebastien-j/LV_groundhog}} as the implementation of the NMT system for all experiments \cite{DBLP:journals/corr/BahdanauCB14,jean15}.
We generally follow the settings and training procedure described by \newcite{DBLP:journals/corr/SennrichHB15}.

For English$\leftrightarrow$German, we report case-sensitive {\sc Bleu} on detokenized text with mteval-v13a.pl for comparison to official WMT and IWSLT results.
For Turkish$\to$English, we report case-sensitive {\sc Bleu} on tokenized text with multi-bleu.perl for comparison to results by \newcite{DBLP:journals/corr/GulcehreFXCBLBS15}.

\newcite{DBLP:journals/corr/GulcehreFXCBLBS15} determine the network vocabulary based on the parallel training data, and replace out-of-vocabulary words with a special UNK symbol.
They remove monolingual sentences with more than 10\% UNK symbols.
In contrast, we represent unseen words as sequences of subword units \cite{DBLP:journals/corr/SennrichHB15}, and can represent any additional training data with the existing network vocabulary that was learned on the parallel data.
In all experiments, the network vocabulary remains fixed.

\subsubsection{English$\leftrightarrow$German}

\begin{table}
\centering
\begin{tabular}{l|r}
dataset & sentences \\
\hline
WMT$_\text{parallel}$ & \num{4200000} \\
WIT$_\text{parallel}$ & \num{200000}\\
\hline
WMT$_\text{mono\_de}$ & \num{160000000} \\
WMT$_\text{synth\_de}$ & \num{3600000} \\
\hline
WMT$_\text{mono\_en}$ & \num{118000000} \\
WMT$_\text{synth\_en}$ & \num{4200000} \\
\end{tabular}
\caption{English$\leftrightarrow$German training data.}
\label{data.en-de}
\end{table}

We use all parallel training data provided by WMT 2015 \cite{bojar-EtAl:2015:WMT}\footnote{\url{http://www.statmt.org/wmt15/}}.
We use the News Crawl corpora as additional training data for the experiments with monolingual data.
The amount of training data is shown in Table \ref{data.en-de}.

Baseline models are trained for a week.
Ensembles are sampled from the last 4 saved models of training (saved at 12h-intervals).
Each model is fine-tuned with fixed embeddings for 12 hours.

For the experiments with synthetic parallel data, we back-translate a random sample of \num{3600000} sentences from the German monolingual data set into English.
The German$\to$English system used for this is the baseline system (\texttt{parallel}).
Translation took about a week on an NVIDIA Titan Black GPU.
For experiments in German$\to$English, we back-translate \num{4200000} monolingual English sentences into German, using the English$\to$German system \texttt{+synthetic}.
Note that we always use single models for back-translation, not ensembles.
We leave it to future work to explore how sensitive NMT training with synthetic data is to the quality of the back-translation.

We tokenize and truecase the training data, and represent rare words via BPE \cite{DBLP:journals/corr/SennrichHB15}.
Specifically, we follow \newcite{DBLP:journals/corr/SennrichHB15} in performing BPE on the joint vocabulary with \num{89500} merge operations.
The network vocabulary size is \num{90000}.

We also perform experiments on the IWSLT 15 test sets to investigate a cross-domain setting.\footnote{\url{http://workshop2015.iwslt.org/}}
The test sets consist of TED talk transcripts.
As in-domain training data, IWSLT provides the WIT$^3$ parallel corpus \cite{cettoloEtAl:EAMT2012}, which also consists of TED talks.

\subsubsection{Turkish$\to$English}

\begin{table}
\centering
\begin{tabular}{l|r}
dataset & sentences \\
\hline
WIT & \num{160000} \\
SETimes & \num{160000} \\
\hline
Gigaword$_\text{mono}$ & \num{177000000} \\
Gigaword$_\text{synth}$ & \num{3200000} \\
\end{tabular}
\caption{Turkish$\to$English training data.}
\label{data.tr-en}
\end{table}

We use data provided for the IWSLT 14 machine translation track \cite{iwslt2014}, namely the WIT$^3$ parallel corpus \cite{cettoloEtAl:EAMT2012}, which consists of TED talks, and the SETimes corpus \cite{tyers10}.\footnote{\url{http://workshop2014.iwslt.org/}}
After removal of sentence pairs which contain empty lines or lines with a length ratio above 9, we retain \num{320000} sentence pairs of training data.
For the experiments with monolingual training data, we use the English LDC Gigaword corpus (Fifth Edition).
The amount of training data is shown in Table \ref{data.tr-en}.
With only \num{320000} sentences of parallel data available for training, this is a much lower-resourced translation setting than English$\leftrightarrow$German.

\newcite{DBLP:journals/corr/GulcehreFXCBLBS15} segment the Turkish text with the morphology tool Zemberek, followed by a disambiguation of the morphological analysis \cite{sak-et-al-cicling-07}, and removal of non-surface tokens produced by the analysis.
We use the same preprocessing\footnote{\url{github.com/orhanf/zemberekMorphTR}}.
For both Turkish and English, we represent rare words (or morphemes in the case of Turkish) as character bigram sequences \cite{DBLP:journals/corr/SennrichHB15}.
The \num{20000} most frequent words (morphemes) are left unsegmented.
The networks have a vocabulary size of \num{23000} symbols.

To obtain a synthetic parallel training set, we back-translate a random sample of \num{3200000} sentences from Gigaword.
We use an English$\to$Turkish NMT system trained with the same settings as the Turkish$\to$English baseline system.

We found overfitting to be a bigger problem than with the larger English$\leftrightarrow$German data set, and follow \newcite{DBLP:journals/corr/GulcehreFXCBLBS15} in using Gaussian noise (stddev 0.01) \cite{NIPS2011_4329}, and dropout on the output layer (p=0.5) \cite{DBLP:journals/corr/abs-1207-0580}.
We also use early stopping, based on {\sc Bleu} measured every three hours on tst2010, which we treat as development set.
For Turkish$\to$English, we use gradient clipping with threshold 5, following \newcite{DBLP:journals/corr/GulcehreFXCBLBS15}, in contrast to the threshold 1 that we use for English$\leftrightarrow$German, following \newcite{jean15}.

\subsection{Results}

\subsubsection{English$\to$German WMT 15}

\begin{table*}
\centering
\begin{tabular}{lc|cccc}
&&\multicolumn{4}{c}{{\sc Bleu}}\\
name & training instances & \multicolumn{2}{c}{newstest2014}& \multicolumn{2}{c}{newstest2015}\\
 &  & single & ens-4 & single & ens-4\\
\hline
\multicolumn{2}{l|}{syntax-based \cite{sennrichhaddow15}} & 22.6 & - & 24.4 & -\\ 
\multicolumn{2}{l|}{Neural MT \cite{jean15b}} & - & - & 22.4 & -\\
\hline
parallel & 37m (parallel) & 19.9 & 20.4 & 22.8 & 23.6 \\ 
+monolingual & 49m (parallel) / 49m (monolingual) & 20.4 & 21.4 & 23.2 & 24.6\\ 
+synthetic & 44m (parallel) / 36m (synthetic) & \textbf{22.7} & \textbf{23.8} & \textbf{25.7} & \textbf{26.5} \\ 
\end{tabular}
\caption{English$\to$German translation performance ({\sc Bleu}) on WMT training/test sets.
Ens-4: ensemble of 4 models. Number of training instances varies due to differences in training time and speed.}
\label{results-ende-wmt}
\end{table*}

Table \ref{results-ende-wmt} shows English$\to$German results with WMT training and test data.
We find that mixing parallel training data with monolingual data with a dummy source side in a ratio of 1-1 improves quality by 0.4--0.5 {\sc Bleu} for the single system, 1 {\sc Bleu} for the ensemble.
We train the system for twice as long as the baseline to provide the training algorithm with a similar amount of parallel training instances.
To ensure that the quality improvement is due to the monolingual training instances, and not just increased training time, we also continued training our baseline system for another week, but saw no improvements in {\sc Bleu}.

Including synthetic data during training is very effective, and yields an improvement over our baseline by 2.8--3.4 {\sc Bleu}.
Our best ensemble system also outperforms a syntax-based baseline \cite{sennrichhaddow15} by 1.2--2.1 {\sc Bleu}.
We also substantially outperform NMT results reported by \newcite{jean15} and \newcite{luong-pham-manning:2015:EMNLP}, who previously reported SOTA result.\footnote{\newcite{luong-pham-manning:2015:EMNLP} report 20.9 {\sc Bleu} (tokenized) on newstest2014 with a single model, and 23.0 {\sc Bleu} with an ensemble of 8 models. Our best single system achieves a \emph{tokenized} {\sc Bleu} (as opposed to untokenized scores reported in Table \ref{results-ende-wmt}) of 23.8, and our ensemble reaches 25.0 {\sc Bleu}.}
We note that the difference is particularly large for single systems, since our ensemble is not as diverse as that of \newcite{luong-pham-manning:2015:EMNLP}, who used 8 independently trained ensemble components, whereas we sampled 4 ensemble components from the same training run.

\subsubsection{English$\to$German IWSLT 15}

\begin{table*}
\centering
\begin{tabular}{ll|lc|ccc}
& name & \multicolumn{2}{c|}{fine-tuning} & \multicolumn{3}{c}{{\sc Bleu}}\\
 & & \multicolumn{1}{c}{data}  & instances & tst2013 & tst2014 & tst2015\\
\hline
 & \multicolumn{3}{l|}{NMT \cite{luong2015} (single model)} & 29.4 & - & -\\
 & \multicolumn{3}{l|}{NMT \cite{luong2015} (ensemble of 8)} & 31.4 & 27.6 & 30.1\\
\hline
1 & parallel & \multicolumn{1}{c}{-} & - & 25.2 & 22.6 & 24.0 \\ 
2 & +synthetic & \multicolumn{1}{c}{-} & - & 26.5 & 23.5 & 25.5 \\ 
\hline
3 & 2+WIT$_\text{mono\_de}$ & WMT$_\text{parallel}$ / WIT$_\text{mono}$ & 200k/200k & 26.6 & 23.6 & 25.4 \\ 
4 & 2+WIT$_\text{synth\_de}$ & WIT$_\text{synth}$ & 200k & 28.2 & 24.4 & 26.7 \\ 
5 & 2+WIT$_\text{parallel}$ & WIT & 200k & \textbf{30.4} & \textbf{25.9} & \textbf{28.4}\\ 
\end{tabular}
\caption{English$\to$German translation performance ({\sc Bleu}) on IWSLT test sets (TED talks). Single models.}
\label{results-ende-iwslt}
\end{table*}

Table \ref{results-ende-iwslt} shows English$\to$German results on IWSLT test sets.
IWSLT test sets consist of TED talks, and are thus very dissimilar from the WMT test sets, which are news texts.
We investigate if monolingual training data is especially valuable if it can be used to adapt a model to a new genre or domain, specifically adapting a system trained on WMT data to translating TED talks.

Systems 1 and 2 correspond to systems in Table \ref{results-ende-wmt}, trained only on WMT data.
System 2, trained on parallel and synthetic WMT data, obtains a {\sc Bleu} score of 25.5 on tst2015.
We observe that even a small amount of fine-tuning\footnote{We leave the word embeddings fixed for fine-tuning.}, i.e.\ continued training of an existing model, on WIT data can adapt a system trained on WMT data to the TED domain.
By back-translating the monolingual WIT corpus (using a German$\to$English system trained on WMT data, i.e.\ without in-domain knowledge), we obtain the synthetic data set WIT$_\text{synth}$.
A single epoch of fine-tuning on WIT$_\text{synth}$ (system 4) results in a {\sc Bleu} score of 26.7 on tst2015, or an improvement of 1.2 {\sc Bleu}.
We observed no improvement from fine-tuning on WIT$_\text{mono}$, the monolingual TED corpus with dummy input (system 3).

These adaptation experiments with monolingual data are slightly artificial in that parallel training data is available.
System 5, which is fine-tuned with the original WIT training data, obtains a {\sc Bleu} of 28.4 on tst2015, which is an improvement of 2.9 {\sc Bleu}.
While it is unsurprising that in-domain parallel data is most valuable, we find it encouraging that NMT domain adaptation with monolingual data is also possible, and effective, since there are settings where only monolingual in-domain data is available.

The best results published on this dataset are by \newcite{luong2015}, obtained with an ensemble of 8 independently trained models.
In a comparison of single-model results, we outperform their model on tst2013 by 1 {\sc Bleu}.

\subsubsection{German$\to$English WMT 15}

\begin{table}
\centering
\begin{tabular}{l|cc}
&\multicolumn{2}{c}{{\sc Bleu}}\\
name & 2014 & 2015\\
\hline
PBSMT \small{\cite{haddow-EtAl:2015:WMT}} & 28.8 & 29.3 \\
\hline
NMT \small{\cite{DBLP:journals/corr/GulcehreFXCBLBS15}} & 23.6 & -\\
+shallow fusion & 23.7 & -\\
+deep fusion & 24.0 & -\\
\hline
parallel & 25.9 & 26.7 \\ 
+synthetic & \textbf{29.5} & \textbf{30.4} \\ 
\hline
+synthetic (ensemble of 4) & 30.8 & 31.6 \\ 
\end{tabular}
\caption{German$\to$English translation performance ({\sc Bleu}) on WMT training/test sets (newstest2014; newstest2015).}
\label{results-deen-wmt}
\end{table}

Results for German$\to$English on the WMT 15 data sets are shown in Table \ref{results-deen-wmt}.
Like for the reverse translation direction, we see substantial improvements (3.6--3.7 {\sc Bleu}) from adding monolingual training data with synthetic source sentences,
which is substantially bigger than the improvement observed with deep fusion \cite{DBLP:journals/corr/GulcehreFXCBLBS15};
our ensemble outperforms the previous state of the art on newstest2015 by 2.3 {\sc Bleu}.

\subsubsection{Turkish$\to$English IWSLT 14}

\begin{table*}
\centering
\begin{tabular}{l|cc|cccc}
name & \multicolumn{2}{c|}{training} & \multicolumn{4}{c}{{\sc Bleu}}\\
 & data & instances & tst2011 & tst2012 & tst2013 & tst2014\\
\hline
\multicolumn{3}{l|}{baseline \cite{DBLP:journals/corr/GulcehreFXCBLBS15}} & 18.4 & 18.8 & 19.9 & 18.7 \\
\multicolumn{3}{l|}{deep fusion \cite{DBLP:journals/corr/GulcehreFXCBLBS15}} & 20.2 & 20.2 & 21.3 & \textbf{20.6}\\
\hline
baseline & parallel & 7.2m & 18.6 & 18.2 & 18.4 & 18.3 \\ 
parallel$_\text{synth}$ & parallel/parallel$_\text{synth}$ & 6m/6m & 19.9 & 20.4 & 20.1 & 20.0 \\ 
\hline
Gigaword$_\text{mono}$ & parallel/Gigaword$_\text{mono}$ & 7.6m/7.6m & 18.8 & 19.6 & 19.4 & 18.2\\ 
Gigaword$_\text{synth}$ & parallel/Gigaword$_\text{synth}$ & 8.4m/8.4m & \textbf{21.2} & \textbf{21.1} & \textbf{21.8} & 20.4\\ 
\end{tabular}
\caption{Turkish$\to$English translation performance (\textit{tokenized} {\sc Bleu}) on IWSLT test sets (TED talks). Single models. Number of training instances varies due to early stopping.}
\label{results-tren}
\end{table*}

Table \ref{results-tren} shows results for Turkish$\to$English.
On average, we see an improvement of 0.6 {\sc Bleu} on the test sets from adding monolingual data with a dummy source side in a 1-1 ratio\footnote{We also experimented with higher ratios of monolingual data, but this led to decreased {\sc Bleu} scores.}, although we note a high variance between different test sets.

With synthetic training data (Gigaword$_\text{synth}$), we outperform the baseline by 2.7 {\sc Bleu} on average,
and also outperform results obtained via shallow or deep fusion by \newcite{DBLP:journals/corr/GulcehreFXCBLBS15} by 0.5 {\sc Bleu} on average.
To compare to what extent synthetic data has a regularization effect, even without novel training data,
we also back-translate the target side of the parallel training text to obtain the training corpus parallel$_\text{synth}$.
Mixing the original parallel corpus with parallel$_\text{synth}$ (ratio 1-1) gives some improvement over the baseline (1.7 {\sc Bleu} on average), but the novel monolingual training data (Gigaword$_\text{mono}$) gives higher improvements, despite being out-of-domain in relation to the test sets.
We speculate that novel in-domain monolingual data would lead to even higher improvements.

\subsubsection{Back-translation Quality for Synthetic Data}

One question that our previous experiments leave open is how the quality of the automatic back-translation affects training with synthetic data.
To investigate this question, we back-translate the same German monolingual corpus with three different German$\to$English systems:

\begin{itemize}
\item with our baseline system and greedy decoding
\item with our baseline system and beam search (beam size 12). This is the same system used for the experiments in Table \ref{results-ende-wmt}.
\item with the German$\to$English system that was itself trained with synthetic data (beam size 12).
\end{itemize}

\begin{table}
\centering
\begin{tabular}{l|c|cc}
&\multicolumn{3}{c}{{\sc Bleu}}\\
& DE$\to$EN & \multicolumn{2}{c}{EN$\to$DE}\\
back-translation & 2015 & 2014 & 2015\\
\hline
none & - & 20.4 & 23.6 \\ 
\hline
parallel (greedy) & 22.3 & 23.2 & 26.0 \\ 
parallel (beam 12) & 25.0 & 23.8 & 26.5 \\ 
synthetic (beam 12) & 28.3 & 23.9 & 26.6 \\ 
\hline
ensemble of 3 & - & 24.2 & 27.0 \\ 
ensemble of 12 & - & 24.7 & 27.6 \\ 
\end{tabular}
\caption{English$\to$German translation performance ({\sc Bleu}) on WMT training/test sets (newstest2014; newstest2015).
Systems differ in how the synthetic training data is obtained. Ensembles of 4 models (unless specified otherwise).}
\label{results-ende-btquality}
\end{table}

{\sc Bleu} scores of the German$\to$English systems, and of the resulting English$\to$German systems that are trained on the different back-translations, are shown in Table \ref{results-ende-btquality}.
The quality of the German$\to$English back-translation differs substantially, with a difference of 6 {\sc Bleu} on newstest2015.
Regarding the English$\to$German systems trained on the different synthetic corpora, we find that the 6 {\sc Bleu} difference in back-translation quality leads to a 0.6--0.7 {\sc Bleu} difference in translation quality.
This is balanced by the fact that we can increase the speed of back-translation by trading off some quality, for instance by reducing beam size, and we leave it to future research to explore how much the amount of synthetic data affects translation quality.

We also show results for an ensemble of 3 models (the best single model of each training run), and 12 models (all 4 models of each training run).
Thanks to the increased diversity of the ensemble components, these ensembles outperform the ensembles of 4 models that were all sampled from the same training run, and we obtain another improvement of 0.8--1.0 {\sc Bleu}.

\subsection{Contrast to Phrase-based SMT}

The back-translation of monolingual target data into the source language to produce synthetic parallel text has been previously explored for phrase-based SMT \cite{Bertoldi_Federico_2009,lambert-EtAl:2011:WMT}.
While our approach is technically similar, synthetic parallel data fulfills novel roles in NMT.

To explore the relative effectiveness of back-translated data for phrase-based SMT and NMT, we train two phrase-based SMT systems with Moses \cite{koehnmoses},
using only WMT$_\text{parallel}$, or both WMT$_\text{parallel}$ and WMT$_\text{synth\_de}$ for training the translation and reordering model.
Both systems contain the same language model, a 5-gram Kneser-Ney model trained on all available WMT data.
We use the baseline features described by \newcite{haddow-EtAl:2015:WMT}.

\begin{table}
\centering
\begin{tabular}{l|cc}
system & \multicolumn{2}{c}{{\sc Bleu}}\\
& WMT & IWSLT\\
\hline
parallel & 20.1 & 21.5 \\ 
+synthetic & 20.8 & 21.6 \\ 
\hline
PBSMT gain & +0.7 & +0.1\\
NMT gain & +2.9 & +1.2\\
\end{tabular}
\caption{Phrase-based SMT results (English$\to$German) on WMT test sets (average of newstest201\{4,5\}), and IWSLT test sets (average of tst201\{3,4,5\}),
and average {\sc Bleu} gain from adding synthetic data for both PBSMT and NMT.}
\label{pbsmt}
\end{table}

Results are shown in Table \ref{pbsmt}.
In phrase-based SMT, we find that the use of back-translated training data has a moderate positive effect on the WMT test sets (+0.7 {\sc Bleu}), but not on the IWSLT test sets.
This is in line with the expectation that the main effect of back-translated data for phrase-based SMT is domain adaptation \cite{Bertoldi_Federico_2009}.
Both the WMT test sets and the News Crawl corpora which we used as monolingual data come from the same source, a web crawl of newspaper articles.\footnote{The WMT test sets are held-out from News Crawl.}
In contrast, News Crawl is out-of-domain for the IWSLT test sets.

In contrast to phrase-based SMT, which can make use of monolingual data via the language model,
NMT has so far not been able to use monolingual data to great effect, and without requiring architectural changes.
We find that the effect of synthetic parallel data is not limited to domain adaptation, and that even out-of-domain synthetic data improves NMT quality, as in our evaluation on IWSLT.
The fact that the synthetic data is more effective on the WMT test sets (+2.9 {\sc Bleu}) than on the IWSLT test sets (+1.2 {\sc Bleu}) supports the hypothesis that domain adaptation
contributes to the effectiveness of adding synthetic data to NMT training.

It is an important finding that back-translated data, which is mainly effective for domain adaptation in phrase-based SMT,
is more generally useful in NMT, and has positive effects that go beyond domain adaptation.
In the next section, we will investigate further reasons for its effectiveness.

\begin{figure}

\begin{tikzpicture}[scale=0.9]
\pgfplotsset{major grid style={style=dotted,color=black!20}}
\begin{axis}[xlabel=training time (training instances $\cdot 10^6$),
    ymin = 2,
    ymax = 8.5,
    xmin = 0,
    xmax = 30,
    ylabel=cross-entropy,
    legend pos = north east,
    legend style={
        font=\scriptsize,
        /tikz/nodes={anchor=west}
        },
    mark size = 0.1,
    ]

    \addplot +[black, mark size=3pt, mark color=black, mark=|, raw gnuplot, solid, line width=0.2ex, id=parallel] gnuplot {plot 'data/exp15.plot' using (($0+1)*0.8):($1/34762/log(2));};
    \addplot +[black, mark size=3pt, mark options={solid}, mark color=black, mark=|, raw gnuplot, dotted, line width=0.15ex, id=parallel-train] gnuplot {plot 'data/exp15-train.plot' using ($1/100000*8):($2/257734/log(2));};

    \addplot +[blue, mark size=1.5pt, mark options={solid, fill=blue}, mark color=blue, mark=triangle*, raw gnuplot, solid, line width=0.2ex, id=synth-p] gnuplot {plot 'data/exp21.plot' using (($0+1)*0.8):($1/34762/log(2));};
    \addplot +[blue, mark size=1.5pt, mark options={solid, fill=blue}, mark color=blue, mark=triangle*, raw gnuplot, dotted, line width=0.15ex, id=synth-p-train] gnuplot {plot 'data/exp21-train.plot' using ($1/100000*8):($2/257734/log(2));};

    \addplot +[red, mark size=1pt, mark options={solid, fill=red}, mark color=red, mark=*, raw gnuplot, solid, line width=0.2ex, id=monolingual] gnuplot {plot 'data/exp16.plot' using (($0+1)*0.8):($1/34762/log(2));};
    \addplot +[red, mark size=1pt, mark options={solid, fill=red}, mark color=red, mark=*, raw gnuplot, dotted, line width=0.15ex, id=monolingual-train] gnuplot {plot 'data/exp16-train.plot' using ($1/100000*8):($2/257734/log(2));};

    \addplot +[orange, mark size=2pt, mark color=orange, mark=x, raw gnuplot, solid, line width=0.2ex, id=synthetic] gnuplot {plot 'data/exp18.plot' using (($0+1)*0.8):($1/34762/log(2));};
    \addplot +[orange, mark size=2pt, mark options={solid}, mark color=orange, mark=x, raw gnuplot, dotted, line width=0.15ex, id=synthetic-train] gnuplot {plot 'data/exp18-train.plot' using ($1/100000*8):($2/257734/log(2));};

    \addlegendentry{parallel (dev)}
    \addlegendentry{parallel (train)}

    \addlegendentry{parallel$_\text{synth}$ (dev)}
    \addlegendentry{parallel$_\text{synth}$ (train)}

    \addlegendentry{Gigaword$_\text{mono}$ (dev)}
    \addlegendentry{Gigaword$_\text{mono}$ (train)}

    \addlegendentry{Gigaword$_\text{synth}$ (dev)}
    \addlegendentry{Gigaword$_\text{synth}$ (train)}

\end{axis}
\end{tikzpicture} 
\caption{Turkish$\to$English training and development set (tst2010) cross-entropy as a function of training time (number of training instances) for different systems.}
\label{cost-tren}
\end{figure}
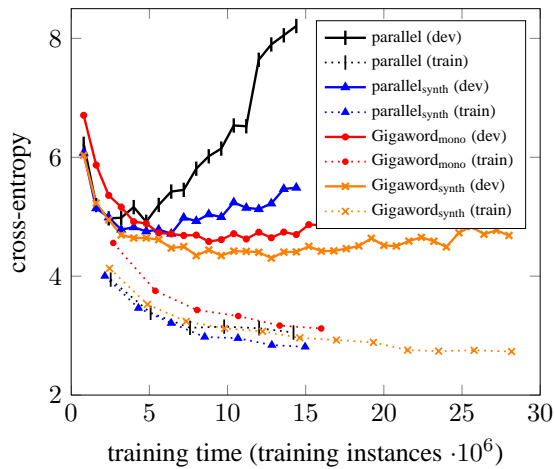

\subsection{Analysis}

\label{sec:analysis}

\begin{figure}

\begin{tikzpicture}[scale=0.9]
\pgfplotsset{major grid style={style=dotted,color=black!20}}
\begin{axis}[xlabel=training time (training instances $\cdot 10^6$),
    ymin = 2,
    ymax = 8.5,
    xmin = 0,
    xmax = 80,
    ylabel=cross-entropy,
    legend pos = north east,
    legend style={
        font=\scriptsize,
        /tikz/nodes={anchor=west}
        },
    mark size = 0.1,
    ]

    \addplot +[black, mark size=3pt, mark color=black, mark=|, raw gnuplot, solid, line width=0.2ex, id=parallel-ende] gnuplot {plot 'data/exp36.plot' using (($1/100000*8)):($2/71803/log(2));};
    \addplot +[black, mark size=3pt, mark options={solid}, mark color=black, mark=|, raw gnuplot, dotted, line width=0.15ex, id=parallel-train-ende] gnuplot {plot 'data/exp36-train.plot' using ($1/100000*8):($2/262029/log(2));};

    \addplot +[orange, mark size=2pt, mark color=orange, mark=x, raw gnuplot, solid, line width=0.2ex, id=synthetic-ende] gnuplot {plot 'data/exp51.plot' using (($1/100000*8)):($2/71803/log(2));};
    \addplot +[orange, mark size=2pt, mark options={solid}, mark color=orange, mark=x, raw gnuplot, dotted, line width=0.15ex, line cap=round, id=synthetic-train-ende] gnuplot {plot 'data/exp51-train.plot' using ($1/100000*8):($2/262029/log(2));};

    \addlegendentry{WMT$_\text{parallel}$ (dev)}
    \addlegendentry{WMT$_\text{parallel}$ (train)}

    \addlegendentry{WMT$_\text{synth}$ (dev)}
    \addlegendentry{WMT$_\text{synth}$ (train)}

\end{axis}
\end{tikzpicture} 
\caption{English$\to$German training and development set (newstest2013) cross-entropy as a function of training time (number of training instances) for different systems.}
\label{cost-ende}
\end{figure}
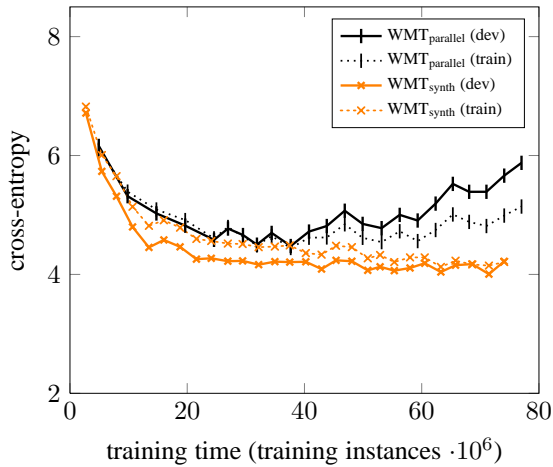

We previously indicated that overfitting is a concern with our baseline system, especially on small data sets of several hundred thousand training sentences, despite the regularization employed.
This overfitting is illustrated in Figure \ref{cost-tren}, which plots training and development set cross-entropy by training time for Turkish$\to$English models.
For comparability, we measure training set cross-entropy for all models on the same random sample of the parallel training set.
We can see that the model trained on only parallel training data quickly overfits, while all three monolingual data sets (parallel$_\text{synth}$, Gigaword$_\text{mono}$, or Gigaword$_\text{synth}$) delay overfitting, and give better perplexity on the development set.
The best development set cross-entropy is reached by Gigaword$_\text{synth}$.

Figure \ref{cost-ende} shows cross-entropy for English$\to$German, comparing the system trained on only parallel data and the system that includes synthetic training data.
Since more training data is available for English$\to$German, there is no indication that overfitting happens during the first 40 million training instances (or 7 days of training);
while both systems obtain comparable training set cross-entropies, the system with synthetic data reaches a lower cross-entropy on the development set.
One explanation for this is the domain effect discussed in the previous section.

A central theoretical expectation is that monolingual target-side data improves the model's fluency, its ability to produce natural target-language sentences.
As a proxy to sentence-level fluency, we investigate word-level fluency, specifically words produced as sequences of subword units, and whether NMT systems trained with additional monolingual data produce more natural words.
For instance, the English$\to$German systems translate the English phrase \emph{civil rights protections} as a single compound, composed of three subword units: \emph{Bürger\pipe rechts\pipe schutzes}\footnote{Subword boundaries are marked with `\pipe'.},
and we analyze how many of these multi-unit words that the translation systems produce are well-formed German words.

We compare the number of words in the system output for the newstest2015 test set which are produced via subword units, and that do not occur in the parallel training corpus.
We also count how many of them are attested in the full monolingual corpus or the reference translation, which we all consider `natural'.
Additionally, the main authors, a native speaker of German, annotated a random subset ($n=100$) of unattested words of each system according to their naturalness\footnote{For the annotation, the words were blinded regarding the system that produced them.},
distinguishing between natural German words (or names) such as \emph{Literatur\pipe klassen} `literature classes', and nonsensical ones such as \emph{*As\pipe best\pipe atten} (a miss-spelling of \emph{Astbestmatten} `asbestos mats').

\begin{table}
\centering
\setlength{\tabcolsep}{3pt}
\begin{tabular}{l|ccc}
system & produced & attested & natural\\
\hline
parallel & 1078 & 53.4\% & 74.9\% \\
+mono & 994 & 61.6\% & 84.6\% \\
+synthetic & 1217 & 56.4\% & 82.5\% \\
\end{tabular}
\caption{Number of words in system output that do not occur in parallel training data ($\text{count}_{\text{ref}}=1168$), and proportion that is attested in data, or natural according to native speaker. English$\to$German; newstest2015; ensemble systems.}
\label{analysis-table}
\end{table}

In the results (Table \ref{analysis-table}), we see that the systems trained with additional monolingual or synthetic data have
a higher proportion of novel words attested in the non-parallel data, and a higher proportion that is deemed natural by our annotator.
This supports our expectation that additional monolingual data improves the (word-level) fluency of the NMT system.

\section{Related Work}

To our knowledge, the integration of monolingual data for pure neural machine translation architectures was first investigated by \cite{DBLP:journals/corr/GulcehreFXCBLBS15},
who train monolingual language models independently, and then integrate them during decoding through rescoring of the beam (\emph{shallow fusion}),
or by adding the recurrent hidden state of the language model to the decoder state of the encoder-decoder network, with an additional controller mechanism that controls the magnitude of the LM signal (\emph{deep fusion}).
In deep fusion, the controller parameters and output parameters are tuned on further parallel training data, but the language model parameters are fixed during the finetuning stage.
\newcite{jean15b} also report on experiments with reranking of NMT output with a 5-gram language model, but improvements are small (between 0.1--0.5 {\sc Bleu}).

The production of synthetic parallel texts bears resemblance to data augmentation techniques used in computer vision,
where datasets are often augmented with rotated, scaled, or otherwise distorted variants of the (limited) training set \cite{Rowley_1996_2678}.

Another similar avenue of research is self-training \cite{McClosky:2006:ESP:1220835.1220855,schwenk08b}.
The main difference is that self-training typically refers to scenario where the training set is enhanced with training instances with artificially produced output labels,
whereas we start with human-produced output (i.e.\ the translation), and artificially produce an input.
We expect that this is more robust towards noise in the automatic translation.
Improving NMT with monolingual source data, following similar work on phrase-based SMT \cite{schwenk08b}, remains possible future work.

Domain adaptation of neural networks via continued training has been shown to be effective for neural language models by \cite{tersarkisov-EtAl:2015:CVSC}, and in work parallel to ours, for neural translation models \cite{luong2015}.
We are the first to show that we can effectively adapt neural translation models with monolingual data.

\section{Conclusion}

In this paper, we propose two simple methods to use monolingual training data during training of NMT systems, with no changes to the network architecture.
Providing training examples with dummy source context was successful to some extent, but we achieve substantial gains in all tasks, and new SOTA results,
via back-translation of monolingual target data into the source language, and treating this synthetic data as additional training data.
We also show that small amounts of in-domain monolingual data, back-translated into the source language, can be effectively used for domain adaptation.
In our analysis, we identified domain adaptation effects, a reduction of overfitting, and improved fluency as reasons for the effectiveness of using monolingual data for training.

While our experiments did make use of monolingual training data, we only used a small random sample of the available data, especially for the experiments with synthetic parallel data.
It is conceivable that larger synthetic data sets, or data sets obtained via data selection, will provide bigger performance benefits.

Because we do not change the neural network architecture to integrate monolingual training data, our approach can be easily applied to other NMT systems.
We expect that the effectiveness of our approach not only varies with the quality of the MT system used for back-translation,
but also depends on the amount (and similarity to the test set) of available parallel and monolingual data, and the extent of overfitting of the baseline model.
Future work will explore the effectiveness of our approach in more settings.

\section*{Acknowledgments}

The research presented in this publication was conducted in cooperation with Samsung Electronics Polska sp.\ z o.o.\ - Samsung R\&D Institute Poland.
This project received funding from the European Union's Horizon 2020 research and innovation programme under grant agreement 645452 (QT21).

\bibliographystyle{acl2016}
\bibliography{../bibliography}

\end{document}